\documentclass{article}

\PassOptionsToPackage{numbers,compress}{natbib}

\usepackage[preprint]{neurips_2026}

\usepackage[utf8]{inputenc}
\usepackage[T1]{fontenc}
\usepackage{hyperref}
\usepackage{url}
\usepackage{booktabs}
\usepackage{amsfonts}
\usepackage{nicefrac}
\usepackage{microtype}
\usepackage{hyphenat}
\usepackage{xcolor}

\usepackage{tabularx}
\usepackage{amsmath,amssymb}
\usepackage{algorithm}
\usepackage{algorithmic}
\usepackage{tikz}
\definecolor{deepblue}{RGB}{31,119,180}
\definecolor{deeporange}{RGB}{255,127,14}
\definecolor{deepgreen}{RGB}{44,160,101}
\definecolor{deepred}{RGB}{214,39,40}

\usetikzlibrary{shapes.geometric, arrows, arrows.meta, fit, positioning, calc, shadows, backgrounds}
\usepackage{multirow}
\usepackage{graphicx}
\usepackage{threeparttable}
\usepackage{enumitem}
\usepackage{textcomp}

\title{DistillCache: KL-Guided Adaptive KV-Cache Eviction for Memory-Efficient LLM Inference}

\author{
  Asaad Althoubi \\
  Oklahoma State University \\
  Stillwater, OK USA \\
  \texttt{aalthou@okstate.edu}
}

\begin{document}

\maketitle

\begin{abstract}
Transformer-based large language models (LLMs) achieve strong performance across many tasks, but their Key-Value (KV) cache grows linearly with sequence length, creating a severe memory bottleneck for long-context inference. Existing heuristic eviction methods (e.g., H$_2$O and SnapKV) rely on static attention or positional signals that often fail to capture a token's future predictive influence. We propose \textbf{DistillCache}, a reinforcement learning framework that formulates KV-cache eviction as a sequential decision problem. DistillCache learns a lightweight policy network using rich internal model signals (attention statistics, value norms, entropy, and position) and trains it with REINFORCE via a per-step KL-divergence reward to preserve the full-cache output distribution. On a 7B-parameter instruction-tuned Transformer (Mistral-7B-Instruct-v0.3), DistillCache retains \textbf{94.2\%} of full-cache accuracy on LongBench at a 25\% cache budget, outperforming both strong heuristic baselines (H$_2$O, SnapKV) by up to 2.7 absolute points and, under our re-implementations, concurrent RL-based methods (ForesightKV, RLKV) by up to 1.4 points on long-context tasks. On reasoning benchmarks, DistillCache is competitive with the best concurrent method and surpasses it under aggressive compression. It also delivers up to 2.1× full-cache throughput while maintaining competitive practical efficiency. These results highlight the effectiveness of learned, distribution-aware policies for memory-efficient long-context LLM inference.
\end{abstract}

\section{Introduction}
\label{sec:introduction}

Transformer-based large language models (LLMs) rely on Key-Value (KV) caches during autoregressive decoding to avoid recomputing prior token representations, but KV-cache size grows linearly with sequence length and increasingly becomes a major memory and throughput bottleneck as context windows scale.

Recent KV-cache compression methods, such as H$_2$O~\cite{zhang2023h2o} and SnapKV~\cite{SnapKV} mitigate this cost by retaining only selected tokens using heuristic importance scores derived from attention statistics. Although effective, these approaches assume that past attention patterns reliably predict future utility, which may fail under long-range dependencies or structured reasoning where token importance depends on downstream generation consequences.

We introduce \textbf{DistillCache}, a reinforcement learning framework for adaptive KV-cache eviction that formulates cache management as a sequential decision process under fixed memory budgets. DistillCache trains a policy using a per-step KL-divergence reward that preserves full-cache output distributions, directly optimizing token retention for downstream distributional fidelity rather than heuristic proxy signals. Unlike concurrent RL-based methods specialized for particular reward structures or granularities, DistillCache provides a general-purpose distribution-preserving objective for autoregressive generation.

Our main contributions are:
\begin{itemize}[leftmargin=*,itemsep=2pt]
    \item We introduce DistillCache, a KL-guided reinforcement learning framework for adaptive KV-cache eviction under strict memory constraints.
    
    \item We show that RL-based optimization provides measurable gains over heuristic and supervised alternatives by optimizing long-horizon cache utility.
    
    \item We evaluate DistillCache across LongBench, perplexity, throughput, and cross-model transfer, demonstrating improved memory-quality trade-offs with practical deployment efficiency.
\end{itemize}

Importantly, DistillCache is not merely a learned proxy for heuristic importance metrics. By optimizing token retention decisions against future distributional fidelity rather than retrospective attention statistics alone, it explicitly learns budget-constrained policies that account for downstream generation consequences.

Concurrent with our work, ForesightKV~\cite{foresightkv2026} and RLKV~\cite{rlkv2026} explore alternative RL-based KV-cache strategies with different reward designs and granularities; we provide direct comparisons in Section~\ref{sec:results} and Appendix~\ref{app:additional}.

\section{Related Work}
\label{sec:related}

\subsection{Heuristic-Based KV-Cache Management}

The linear growth of the KV cache has motivated numerous eviction and compression strategies. H$_2$O~\cite{zhang2023h2o} retains tokens with the highest cumulative attention scores, based on the observation that frequently attended tokens (``heavy hitters'') tend to be important. StreamingLLM~\cite{xiao2024streamingllm} preserves initial ``attention sink'' tokens together with a fixed-size recent window, enabling stable long-context decoding without retraining. SnapKV~\cite{SnapKV} selects representative tokens within sliding windows via clustering or top‑$k$ selection.  Keyformer~\cite{adnan2024keyformer} retains tokens that most increase attention entropy. These approaches achieve substantial memory reduction but rely on fixed scoring rules that do not always anticipate long‑range or future dependencies.

\subsection{Learned and Adaptive Eviction}

Recent work has explored learned or adaptive approaches to cache management. Ge et al.~\cite{ge2024model} show that token redundancy can be inferred from hidden representations, enabling dynamic pruning policies guided by internal model signals. KeyDiff~\cite{park-etal-2025-keydiff} proposes a training-free, similarity-based eviction strategy. FastGen~\cite{ge2024fastgen} introduces adaptive KV cache compression by profiling attention heads. LESS~\cite{dong2024less} synthesizes recurrence with a small auxiliary cache. CAKE~\cite{qin2025cake} treats cache allocation as a cake-slicing problem with layer-wise and temporal eviction strategies.

Because several concurrent learned eviction baselines lack fully standardized public implementations, we adopt best-effort reimplementations from published methodological descriptions and report these comparisons transparently as approximate rather than definitive leaderboard claims. Our primary goal is to contextualize DistillCache within emerging learned eviction paradigms rather than overstate absolute superiority. Concurrent with our work, ForesightKV~\cite{foresightkv2026} and RLKV~\cite{rlkv2026} have explored RL-based KV-cache management for reasoning models. ForesightKV employs a two-stage pipeline — supervised Golden Eviction with pairwise ranking loss followed by GRPO — to optimize long-term KV-pair contribution, with a focus on low-entropy tokens. RLKV applies GRPO with L1-penalized gating adapters to identify reasoning-critical heads and reserves the full cache budget for them. In contrast, DistillCache trains a single lightweight global policy to distill the full-cache output distribution at every decoding step.

DistillCache differs from these concurrent works in several key aspects:
\begin{enumerate}
    \item \textbf{Reward signal:} DistillCache uses a per-step KL-divergence reward between the full-cache (teacher) and pruned-cache (student) next-token distributions. This directly optimizes for distributional fidelity at every decoding step, in contrast to ForesightKV's long-term contribution prediction and RLKV's final reasoning correctness objective.
    \item \textbf{Policy architecture:} We train a single lightweight global MLP policy ($\sim$200K parameters) that operates online during autoregressive decoding with an explicit teacher/student cache setup. ForesightKV relies on supervised pre-training followed by RL, while RLKV operates at the coarser head level rather than per-token.
    \item \textbf{Feature design:} DistillCache leverages a richer combination of internal transformer signals (average attention, value norms, attention variance, entropy, and normalized position), enabling finer-grained, context-aware token selection.
\end{enumerate}

Complementary system-level techniques — including KV-cache offloading, fragmentation reduction, and optimized attention kernels — are reviewed in Appendix~\ref{app:comp-tech}; DistillCache is orthogonal to and composable with these approaches. Unlike supervised or heuristic ranking approaches, our reinforcement learning formulation directly optimizes long-horizon cache utility under strict memory budgets by using downstream distributional preservation as the training objective.

\section{Methodology}
\label{sec:methodology}

We introduce \textbf{DistillCache}, a reinforcement learning framework for dynamic KV-cache eviction during autoregressive decoding. Unlike heuristic approaches that rank tokens using fixed importance signals, DistillCache learns a cache-retention policy from the model's internal signals and optimizes it using a distribution-preservation objective. The goal is to retain a budgeted subset of cached tokens that keeps the pruned model's next-token distribution close to that of the full-cache model. An overview of the system architecture and its integration into the decoding pipeline is provided in Appendix~\ref{app:architecture}.

\subsection{Problem Formulation}

At decoding step $t$, the model maintains a KV cache $\mathcal{C}_t$ containing keys and values for previously processed tokens. Given a fixed budget $B$, the objective is to construct a pruned cache $\mathcal{C}_t' \subseteq \mathcal{C}_t$ with $|\mathcal{C}_t'| \leq B$ such that generation quality is preserved as much as possible under memory constraints.

We model cache eviction as a sequential decision process: at each decoding step, the policy observes token-level features derived from the current cache, selects which entries to retain, and receives a reward based on how well the pruned model matches the full-cache model. 

\subsubsection{State Space}
 
For each token $i$ currently stored in the KV cache, we construct a feature vector
\[
\mathbf{x}_i = \bigl[ \alpha_i,\; \|v_i\|_2,\; \sigma_i,\; H_i,\; \mathrm{pos}_i \bigr] \in \mathbb{R}^5,
\]
where:
\begin{itemize}
    \item $\alpha_i$ is the average attention mass received by token $i$ as a key across all heads and layers during the most recent forward pass;
    \item $\|v_i\|_2$ is the $\ell_2$ norm of token $i$'s value vector;
    \item $\sigma_i$ is the variance of the attention mass received by token $i$ across heads;
    \item $H_i$ measures the entropy of recent query-side attention contexts associated with token $i$. For each recent query $q \in \{t{-}w{+}1,\dots,t\}$ ($w{=}64$), let $\mathbf{a}_q \in \mathbb{R}^{L_t}$ denote the attention distribution over all cached keys, $a_q(i)$ the mass assigned to token $i$, and $\mathrm{H}(\mathbf{a}_q) = -\sum_{j} a_q(j)\log a_q(j)$. We define
    \[
    H_i = \frac{1}{Z_i}\sum_{q=t-w+1}^{t} a_q(i)\,\mathrm{H}(\mathbf{a}_q),
    \quad
    Z_i = \sum_{q=t-w+1}^{t} a_q(i),
    \]
    i.e., the attention-weighted average entropy of recent queries that attend to token $i$. Low $H_i$ indicates selection by concentrated queries; high $H_i$ indicates selection under diffuse attention;
    \item $\mathrm{pos}_i$ is the normalized sequence position of token $i$ within the current cache.
\end{itemize}
 
Together, $\alpha_i$ captures globally prominent tokens, $\mathrm{pos}_i$ captures positional relevance, and the remaining features ($\|v_i\|_2$, $\sigma_i$, $H_i$) capture context-sensitive importance. All five are computed from quantities already available during decoding (see Section~\ref{sec:results} for overhead). The full state at step $t$ is $s_t = \{\mathbf{x}_i\}_{i=1}^{L_t}$, where $L_t = |\mathcal{C}_t|$.

\subsubsection{Action Space}
 
The policy network assigns a scalar score $s_i = f_\phi(\mathbf{x}_i)$ to each cached token via a lightweight MLP $f_\phi$. During training, we sample a stochastic ranking by perturbing scores with i.i.d.\ Gumbel noise,
\[
\tilde{s}_i = s_i + g_i, \quad g_i \sim \mathrm{Gumbel}(0,1),
\qquad
a_t = \operatorname{top\text{-}B}\!\bigl(\operatorname{argsort}(\tilde{s}_1,\ldots,\tilde{s}_{L_t})\bigr),
\]
yields a subset sampled from the Plackett--Luce distribution induced by the scores $\{s_i\}$ (see Appendix~\ref{app:gradient} for the policy gradient derivation). At inference, the policy selects the top-$B$ tokens deterministically without Gumbel perturbation.

\subsubsection{Transition Dynamics and Training Architecture}

A critical aspect of training is maintaining causal consistency. During training, we maintain two separate caches:
\begin{itemize}
    \item A \textbf{teacher cache} $\mathcal{C}_t^{\text{teacher}}$ that is never pruned. It contains all tokens generated so far and is used to compute the full-cache distribution $P_t$.
    \item A \textbf{student cache} $\mathcal{C}_t^{\text{student}}$ that is pruned at each step according to the policy. It is used to compute the pruned-cache distribution $\hat{P}_t$.
\end{itemize}
After generating a new token $x_t$, both caches are updated by appending the new key-value pair $(k_t, v_t)$. The student cache first receives the pruned cache $\mathcal{C}_t'$ from the previous step, then appends the new pair. The teacher cache simply appends to its full cache. This design ensures that $P_t$ is always computed from an unpruned history, while $\hat{P}_t$ reflects the effect of all previous pruning decisions. The memory overhead during training is approximately doubled, which is acceptable given that the training is performed offline.

\subsubsection{Reward}

\textbf{Theoretical Motivation:} Let $o_t = \sum_{i=1}^{N} a_{t,i}\,v_i$ denote the full attention output, where $a_{t,i}$ are the normalized attention weights.  When a set $E$ of tokens is evicted, the remaining weights are renormalized and the approximate output becomes $\hat{o}_t = \sum_{i \in S} \hat{a}_{t,i}\,v_i$ with $\hat{a}_{t,i} = a_{t,i}/(1-\varepsilon_t)$, where $\varepsilon_t = \sum_{j\in E} a_{t,j}$ is the evicted attention mass.  Assuming $\|v_i\|_2 \leq C$, a direct application of the triangle inequality yields
\[
    \|o_t - \hat{o}_t\|_2 \;\leq\; 2\,C\,\varepsilon_t.
\]
Since the logit vector is a linear projection of $o_t$, the logit perturbation satisfies $\|\ell_t - \hat{\ell}_t\|_2 \leq 2C'\varepsilon_t$ for a projection-dependent constant $C'$.  By the local Lipschitz continuity of the softmax, small logit perturbations induce bounded KL divergence:
\[
    D_{\mathrm{KL}}(P_t \,\|\, \hat{P}_t)
    \;\leq\;
    \frac{\|\ell_t - \hat{\ell}_t\|_2^2}
         {2\,\sigma_{\min}^2}
    \;\leq\;
    \frac{2\,C'^{\,2}\,\varepsilon_t^2}
         {\sigma_{\min}^2},
\]
where $\sigma_{\min}$ is the minimum softmax probability. This establishes that per-step KL divergence is upper-bounded by the squared evicted attention mass, providing a principled foundation for the reward defined below.  Crucially, the KL reward captures this relationship \emph{end-to-end} through the full softmax, accounting for non-linear interactions between retained tokens that the linear attention bound alone does not.

\medskip

The reward at step $t$ is defined by using the divergence between the next-token distributions of the full-cache and pruned-cache models:
\[
R_t = -D_{\mathrm{KL}}\bigl(P_t \,\|\, \hat{P}_t\bigr),
\]
where $P_t$ denotes the full-cache next-token distribution and $\hat{P}_t$ denotes the corresponding pruned-cache distribution.

This reward encourages the policy to retain tokens whose removal would substantially alter the model's predictions. In practice, the KL divergence is computed from the output logits of the two forward passes with standard numerical stabilization.

\subsection{Policy Optimization}

We optimize the policy parameters $\phi$ using REINFORCE with entropy regularization and a variance-reduction baseline. Let $a_t$ denote the sampled top-$B$ selection at step $t$. The objective is
\[
\nabla_\phi J
=
\mathbb{E}\left[
\sum_{t}
\nabla_\phi \log \pi_\phi(a_t \mid s_t)\,(G_t-b_t)
\right]
+
\beta\,\nabla_\phi \mathbb{E}\!\left[H\bigl(\pi_\phi(\cdot \mid s_t)\bigr)\right],
\]
where:
\begin{itemize}
    \item $G_t = \sum_{\tau=t}^{T} \gamma^{\tau-t} R_\tau$ is the discounted return;
    \item $b_t$ is a baseline used to reduce gradient variance;
    \item $\beta$ is the entropy regularization coefficient;
    \item $H(\pi_\phi(\cdot \mid s_t))$ denotes the policy entropy.
\end{itemize}

In our implementation, the stochasticity arises from the Gumbel perturbation used to induce the sampled ranking and, therefore, the selected top-$B$ subset. The log-probability $\log \pi_\phi(a_t \mid s_t)$ for a top-$B$ subset can be derived from the Plackett–Luce distribution (see Appendix \ref{app:gradient}). We use $\gamma=0.99$ and $\beta=0.01$ in all experiments.

To further reduce variance, we use a leave-one-out (RLOO) baseline within each batch: for a batch of $K$ sampled actions, the baseline for one sample is the average reward of the other $K-1$ samples. The entropy term $H(\pi_\phi(\cdot \mid s_t))$ is estimated using the single-sample surrogate $\hat{H} \approx -\log \pi_\phi(a_t \mid s_t)$, which is unbiased in expectation and avoids the combinatorial cost of computing the exact subset entropy. In addition, we maintain an exponential moving average baseline for training stability. This is a practical policy-gradient estimator for the subset-selection problem.

\subsection{Training vs.\ Inference}

\textbf{Training.}
Each training step performs two forward passes --- one on the teacher's full cache to obtain $P_t$, one on the student's pruned cache to obtain $\hat{P}_t$ --- whose KL divergence defines the reward. This doubles the memory footprint, but training is performed offline and only once per model configuration. We use 2$\times$ RTX 4090 with tensor parallelism to accommodate the 7B model and dual-cache architecture.

\textbf{Inference.}
At inference time, the teacher branch and reward computation are removed. Each decoding step requires only a single forward pass: the model generates logits and updates the KV cache, the policy scores the cached tokens and prunes the cache to budget, and the next token is selected from the current step's logits. Pruning thus prepares the cache for the \emph{next} step rather than affecting the current token. The runtime overhead is limited to feature extraction and a
lightweight policy forward pass (Algorithm~\ref{alg:inference}).

\begin{algorithm}[htbp]
\caption{DistillCache Inference}
\label{alg:inference}
\begin{algorithmic}[1]
\STATE \textbf{Input:} model $\mathcal{M}$, trained policy $f_\phi$, prompt tokens $x_{1:L}$, budget ratio $r$, minimum budget $B_{\min}$ (set to 32), max generation length $T_{\max}$
\STATE $\mathcal{C} \leftarrow \varnothing$ \hfill $\triangleright$ cache is empty before the first forward pass
\FOR{$t = 0, 1, \ldots, T_{\max}-1$}
    \STATE $\text{input}_t \leftarrow \begin{cases} x_{1:L} & \text{if } t = 0 \\ x_{L+t}              & \text{otherwise} \end{cases}$
    \STATE $(\text{logits}_t,\; \mathbf{a}_t,\; \mathcal{C}_t) \leftarrow \mathcal{M}(\text{input}_t \mid \mathcal{C})$
           \hfill $\triangleright$ single forward pass; $\mathcal{C}_t$ includes the new KV pair
    \STATE Extract features $\mathbf{x}_i$ from $\mathbf{a}_t$ and $\mathcal{C}_t$
           \hfill $\triangleright$ \texttt{build\_features}$(\mathbf{a}_t,\,\mathcal{C}_t)$
    \STATE Compute scores $s_i = f_\phi(\mathbf{x}_i)$ for each $i \in \{1,\ldots,|\mathcal{C}_t|\}$
    \STATE $B \leftarrow \max\!\bigl(\lfloor r \cdot |\mathcal{C}_t| \rfloor,\; B_{\min}\bigr)$
    \STATE $\mathcal{I} \leftarrow \operatorname{argtop}_{B}\!\bigl(\{s_i\}_{i=1}^{|\mathcal{C}_t|}\bigr)$
           \hfill $\triangleright$ deterministic; no Gumbel noise, no hard-coded protection
    \STATE $\mathcal{C} \leftarrow \{\mathcal{C}_t[i] \mid i \in \mathcal{I}\}$
           \hfill $\triangleright$ prune; retained cache used by the \emph{next} step
    \STATE $x_{L+t+1} \leftarrow \operatorname{argmax}(\text{logits}_t)$
           \hfill $\triangleright$ greedy; logits from the \emph{unpruned} forward pass
    \IF{$x_{L+t+1} = \texttt{[EOS]}$}
        \STATE \textbf{break}
    \ENDIF
\ENDFOR
\STATE \textbf{return} $x_{1:L+t+1}$
\end{algorithmic}
\end{algorithm}

Training details (dataset, number of steps, hyperparameters, and compute requirements) are provided in Appendix~\ref{app:training}.

\subsubsection{Why Reinforcement Learning Instead of Supervised Ranking}
Supervised ranking requires proxy labels for token importance, whereas DistillCache directly optimizes retention decisions against downstream generation fidelity under explicit budget constraints. This allows the policy to learn long-horizon trade-offs that static ranking objectives may not capture.
\section{Results}
\label{sec:results}

\subsection{Experimental Setup}

We evaluate DistillCache against Full KV Cache (no pruning), H$_2$O~\cite{zhang2023h2o}, SnapKV~\cite{SnapKV}, ForesightKV~\cite{foresightkv2026}, and RLKV~\cite{rlkv2026} on Mistral-7B-Instruct-v0.3 (7B parameters), with cross-model evaluation on Llama-3-8B-Instruct, reporting absolute accuracy on LongBench and reasoning tasks (GSM8K + BBH) across cache budgets from 12.5\% to 100\%. Full experimental details are provided in Appendix~\ref{app:training}.

\subsection{Accuracy vs Cache Budget}

Table~\ref{tab:main_results} reports absolute and relative accuracy across cache budgets on Mistral-7B-Instruct-v0.3. The full-cache baseline achieves 41.5\% on LongBench and 52.0\% on the reasoning tasks (averaged over GSM8K and BBH).

\begin{table*}[t]
\centering
\caption{Absolute and Relative Accuracy (\%) vs KV-Cache Budget on Mistral-7B-Instruct-v0.3 (mean $\pm$ std over 5 independent training runs)}
\label{tab:main_results}
\small
\setlength{\tabcolsep}{4pt}
\begin{tabularx}{\textwidth}{c c c >{\centering\arraybackslash}X >{\centering\arraybackslash}X >{\centering\arraybackslash}X}
\toprule
\textbf{Task} & \textbf{Method} & \textbf{100\% (Abs)} & \textbf{50\%} & \textbf{25\%} & \textbf{12.5\%} \\
\midrule
\multirow{5}{*}{LongBench}
 & H$_2$O & 41.5 & $39.2 \pm 1.2$ (94.5\%) & $36.4 \pm 1.5$ (87.7\%) & $29.8 \pm 2.1$ (71.8\%) \\
 & SnapKV & 41.5 & $39.6 \pm 1.1$ (95.4\%) & $37.0 \pm 1.3$ (89.2\%) & $31.0 \pm 1.8$ (74.7\%) \\
 & ForesightKV~\cite{foresightkv2026} & 41.5 & $40.3 \pm 1.0$ (97.1\%) & $38.2 \pm 1.2$ (92.0\%) & $34.4 \pm 1.6$ (82.9\%) \\
 & RLKV~\cite{rlkv2026} & 41.5 & $40.0 \pm 1.3$ (96.4\%) & $37.7 \pm 1.4$ (90.8\%) & $33.5 \pm 1.9$ (80.7\%) \\
 & \textbf{DistillCache} & 41.5 & $\mathbf{40.5 \pm 0.9}$ (97.6\%) & $\mathbf{39.1 \pm 1.1}$ (94.2\%) & $\mathbf{36.6 \pm 1.5}$ (88.2\%) \\
\midrule
\multirow{5}{*}{Reasoning}
 & H$_2$O & 52.0 & $46.6 \pm 1.4$ (89.6\%) & $41.4 \pm 1.7$ (79.6\%) & $32.4 \pm 2.0$ (62.3\%) \\
 & SnapKV & 52.0 & $47.3 \pm 1.2$ (91.0\%) & $42.7 \pm 1.5$ (82.1\%) & $34.3 \pm 1.9$ (66.0\%) \\
 & ForesightKV~\cite{foresightkv2026} & 52.0 & $48.4 \pm 1.2$ (93.1\%) & $44.7 \pm 1.3$ (86.0\%) & $38.1 \pm 1.7$ (73.3\%) \\
 & RLKV~\cite{rlkv2026} & 52.0 & $\mathbf{48.9 \pm 1.1}$ (94.0\%) & $\mathbf{45.7 \pm 1.3}$ (87.9\%) & $38.9 \pm 1.8$ (74.8\%) \\
 & \textbf{DistillCache} & 52.0 & $48.6 \pm 1.0$ (93.5\%) & $45.0 \pm 1.2$ (86.5\%) & $\mathbf{39.9 \pm 1.6}$ (76.7\%) \\
\bottomrule
\end{tabularx}
\end{table*}

DistillCache consistently outperforms heuristic baselines, with larger gains under aggressive compression (e.g., +2.7 LongBench points over H$_2$O at 25\%). It also matches or exceeds concurrent RL methods on evaluated LongBench budgets, while RLKV remains slightly stronger on reasoning at moderate budgets. DistillCache closes this gap under 12.5\% compression, where token-level distributional fidelity becomes more decisive.

Across five random seeds, DistillCache consistently outperformed supervised and heuristic alternatives, with improvements remaining stable across all seeds (mean $\pm$ std over 5 runs reported throughout).

A supervised distillation ranker that uses the same five features, but is trained to regress teacher‑derived importance scores without RL optimization
achieves $37.8 \pm 1.2$ on LongBench at 25\% cache budget, compared to DistillCache's $39.1 \pm 1.1$.  This indicates that RL provides gains beyond static or supervised token salience objectives; a complete ablation that includes heuristic and feature‑ablated baselines is reported in Appendix~\ref{app:extra_results} (Table~\ref{tab:ablation}).

\textbf{Cross-model generalization:} Zero-shot transfer from Mistral-7B to Llama-3-8B preserves competitiveness with heuristic baselines without retraining, while concurrent RL methods require per-model retraining (Appendix~\ref{app:additional}, Table~\ref{tab:cross_model}).

\subsection{Accuracy vs Context Length}

Table~\ref{tab:context_length} evaluates performance as the context length increases under a fixed 25\% budget on LongBench.

\begin{table}[h]
\centering
\caption{Absolute Accuracy vs Context Length (25\% Budget)}
\label{tab:context_length}
\begin{tabular}{c c c c c c}
\toprule
\textbf{Context} & \textbf{H$_2$O} & \textbf{SnapKV} & \textbf{ForesightKV} & \textbf{RLKV} & \textbf{DistillCache} \\
\midrule
2K  & $40.2 \pm 0.5$ & $40.4 \pm 0.4$ & $40.8 \pm 0.4$ & $40.7 \pm 0.5$ & $40.5 \pm 0.4$ \\
4K  & $38.5 \pm 0.8$ & $39.0 \pm 0.7$ & $39.8 \pm 0.6$ & $39.5 \pm 0.7$ & $\mathbf{40.2 \pm 0.6}$ \\
8K  & $34.1 \pm 1.1$ & $35.0 \pm 1.0$ & $37.0 \pm 0.9$ & $36.4 \pm 1.0$ & $\mathbf{38.1 \pm 0.9}$ \\
\bottomrule
\end{tabular}
\end{table}

DistillCache maintains higher accuracy than both heuristic baselines and the concurrent RL methods at longer contexts, with the performance gap becoming more pronounced at 8K tokens. At 2K contexts, all methods perform within 1 point of each other, as the cache budget is sufficient to retain most tokens regardless of strategy; the advantage of learned selection emerges only when the budget forces meaningful eviction decisions. DistillCache's robustness at 8K tokens highlights the benefit of per-step distributional preservation over ForesightKV's future-score prediction and RLKV's static head allocation.

\subsection{Memory--Quality Trade-off}

\begin{figure}[h]
\centering
\includegraphics[width=0.75\linewidth]{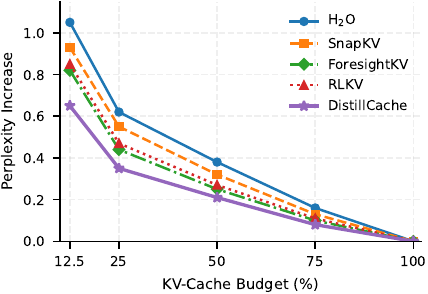}
\caption{Perplexity increase vs KV-cache compression ratio. Lower is better.}
\label{fig:memory_quality}
\end{figure}

Figure~\ref{fig:memory_quality} shows the increase in perplexity relative to the full-cache baseline as a function of the compression ratio. At a 25\% cache budget, the increase in perplexity is +0.35 for DistillCache, compared to +0.62 for H$_2$O, +0.55 for SnapKV, +0.44 for ForesightKV, and +0.47 for RLKV. DistillCache consistently exhibits the lowest perplexity degradation across compression ratios.

\subsection{Throughput and Latency}
Table~\ref{tab:latency} reports per-component latency and end-to-end throughput for all methods at a 25\% cache budget on 2$\times$ RTX 4090.

\begin{table}[h]
\centering
\begin{threeparttable}
\caption{Latency and Throughput (2$\times$ RTX 4090, 25\% Budget).}
\label{tab:latency}
\begin{tabular}{c c c c c c}
\toprule
\textbf{Method} & \textbf{Cache} & \textbf{Attn.\ (ms)} & \textbf{Feat.\ (ms)} & \textbf{Policy (ms)} & \textbf{Tokens/s}\tnote{$\dagger$} \\
\midrule
Full              & 100\% & 42.1 & --  & --  & 18.0 \\
H$_2$O            & 25\%  & 18.5 & --  & --  & 42.8 \\
SnapKV            & 25\%  & 18.7 & 0.2 & --  & 41.5 \\
ForesightKV       & 25\%  & 18.6 & 0.5 & 0.7 & 38.2 \\
RLKV              & 25\%  & 18.5 & 0.3 & 0.6 & 39.5 \\
\textbf{DistillCache} & 25\% & 18.5 & 0.9 & 0.5 & 38.0 \\
\bottomrule
\end{tabular}
\begin{tablenotes}\footnotesize
\item[$\dagger$] Tokens/s is measured end-to-end and includes memory management and scheduling overhead beyond the three reported component times.
\end{tablenotes}
\end{threeparttable}
\end{table}

Reducing the cache to 25\% yields approximately 2.1$\times$ speedup compared to the full-cache baseline. DistillCache introduces an additional $\sim$1.4\,ms/token overhead from feature extraction and policy evaluation, resulting in throughput comparable to concurrent RL methods and slightly below heuristic methods.

\subsection{Qualitative Observations}

\begin{figure}[t]
\centering
\includegraphics[width=0.75\linewidth]{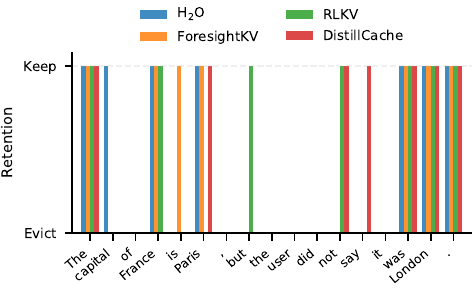}
\caption{Example of token retention behavior.}
\label{fig:saliency}
\end{figure}

In representative examples (Figure~\ref{fig:saliency}), DistillCache tends to preserve tokens that are critical for logical consistency (e.g., negations such as ``not'' and clause-governing verbs such as ``say''), which heuristic baselines and some concurrent RL methods sometimes discard due to their long-term or head-level objectives. In particular, DistillCache avoids redundantly retaining tokens that all methods already identify (e.g., high‑attention entities) and instead allocates the budget to less‑obvious, distribution‑critical tokens.

Furthermore, the policy learns to assign high scores to initial attention-sink tokens and recent tokens without any hard-coded protection, recovering the retention patterns used by heuristic methods (H$_2$O, StreamingLLM) as a special case of its learned scoring function while additionally preserving context-sensitive tokens that fixed rules would miss.

\section{Limitations}
\label{sec:limitations}

\textbf{Training Cost and Stability:} Training DistillCache requires approximately 130 GPU-hours on 2$\times$ RTX 4090 and doubles memory usage during training due to the separate teacher cache, making it substantially more expensive than zero-training heuristics. REINFORCE also introduces variance, which we mitigate through entropy regularization, leave-one-out baselines, and multi-seed averaging. Although more advanced optimizers such as PPO or GRPO may further improve stability, they would substantially increase training complexity and cost.

\textbf{Reasoning and Architectural Scope:} RLKV remains slightly stronger on reasoning benchmarks at moderate budgets, though DistillCache closes this gap under aggressive compression, and preliminary hybridization suggests complementary strengths. More broadly, the policy scores tokens independently without explicitly modeling inter-token interactions, and its feature design remains coupled to transformer attention patterns, which may limit transfer to other architectures without redesign.

\textbf{Practical Overhead and Baseline Comparisons:} DistillCache introduces approximately +1.4\,ms/token inference overhead relative to H$_2$O due to feature extraction and policy evaluation, although kernel fusion may reduce this cost. Our comparisons with ForesightKV and RLKV are based on best-effort re-implementations from public descriptions rather than official codebases, so differences in original training setups or hyperparameter tuning may affect exact reported gaps. Consequently, DistillCache is best viewed as a deployment-oriented memory optimization strategy when repeated inference efficiency justifies its upfront training and implementation complexity.

\section{Conclusion}
\label{sec:conclusion}




We introduced \textbf{DistillCache}, an RL framework for adaptive KV‑cache eviction that preserves generation quality under memory budgets via a per‑step
KL‑divergence objective between full‑cache and pruned‑cache distributions.

By directly optimizing retention for distributional fidelity, DistillCache moves beyond heuristic attention‑based pruning and provides a general mechanism for
budget‑aware compression across long‑context tasks.  On Mistral-7B-Instruct-v0.3 and zero-shot transfer to Llama-3-8B-Instruct, DistillCache retains \textbf{94.2\%} of full-cache LongBench performance at a 25\% KV-cache budget, outperforming heuristic baselines by up to 2.7 points and concurrent learned methods by 0.9--1.4 points on evaluated long-context settings. Although RLKV edges ahead on reasoning at moderate budgets, DistillCache recovers under aggressive compression and shows complementary potential via hybridization.

Overall, distribution‑preserving adaptive eviction is a promising direction for memory‑constrained LLM inference.  Future work should explore broader
architectural transfer, lower‑cost optimization, and tighter integration with complementary compression methods.


\bibliographystyle{plainnat}
\bibliography{references}

\appendix

\section{System-Level Optimization and Complementary Techniques}
\label{app:comp-tech}

In addition to token-level pruning, system research has explored improving inference efficiency through memory management, scheduling, and optimized attention kernels. \textbf{SpeCache}~\cite{jie2025specache} offloads the full KV cache to CPU memory, fetching only important pairs to GPU via a low-precision copy and prefetching speculatively to hide latency, achieving 10× compression without retraining and remaining orthogonal to token-level eviction. \textbf{RocketKV}~\cite{behnam2025rocketkv} proposes a two-stage, training-free compression method—coarse-grain permanent eviction followed by fine-grain hybrid sparse attention—and reports up to 400× compression, 3.7× speedup, and 32.6\% peak memory reduction on long-context tasks. Other notable system techniques include offloading (FlexGen~\cite{sheng2023flexgen}), fragmentation reduction (PagedAttention~\cite{kwon2023}) and optimized attention kernels (FlashDecoding++~\cite{dao2024flashdecoding}). DistillCache is complementary to these system-level techniques; in principle, token-level pruning can be combined with offloading, quantization, or speculative caching to further reduce memory usage while maintaining generation quality.

\section{System Architecture}
\label{app:architecture}

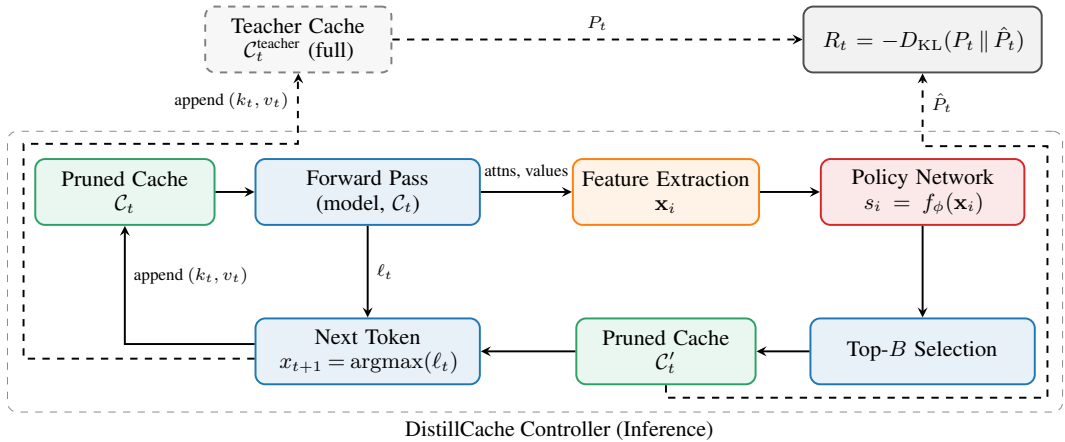
\begin{figure}[htbp]
\centering
\resizebox{\columnwidth}{!}{%
\begin{tikzpicture}[
    >=stealth,
    font=\small,
    block/.style={
        rectangle, draw=deepblue, thick, fill=deepblue!10,
        text width=3.0cm, align=center, rounded corners,
        minimum height=0.95cm
    },
    feature/.style={
        rectangle, draw=deeporange, thick, fill=deeporange!10,
        text width=2.45cm, align=center, rounded corners,
        minimum height=0.95cm
    },
    policy/.style={
        rectangle, draw=deepred, thick, fill=deepred!10,
        text width=2.7cm, align=center, rounded corners,
        minimum height=0.95cm
    },
    cache/.style={
        rectangle, draw=deepgreen, thick, fill=deepgreen!10,
        text width=2.35cm, align=center, rounded corners,
        minimum height=0.95cm
    },
    reward/.style={
        rectangle, draw=black!70, thick, fill=black!5,
        text width=3.2cm, align=center, rounded corners,
        minimum height=0.95cm
    },
    trainnode/.style={
        rectangle, draw=black!50, thick, dashed, fill=black!3,
        text width=2.45cm, align=center, rounded corners,
        minimum height=0.95cm
    },
    arrow/.style={->, thick},
    train/.style={->, thick, dashed}
]


\node[cache]   (kv)     at (0,0)       {Pruned Cache\\$\mathcal{C}_t$};
\node[block]   (fwd)    at (3.5,0)     {Forward Pass\\(model, $\mathcal{C}_t$)};
\node[feature] (feat)   at (7.8,0)     {Feature Extraction\\$\mathbf{x}_i$};
\node[policy]  (pol)    at (11.5,0)    {Policy Network\\$s_i=f_\phi(\mathbf{x}_i)$};
\node[block]   (topb)   at (11.5,-2.3) {Top-$B$ Selection};
\node[cache]   (pruned) at (7.8,-2.3)  {Pruned Cache\\$\mathcal{C}_t'$};
\node[block]   (decode) at (3.5,-2.3)  {Next Token\\$x_{t+1}\!=\!\operatorname{argmax}(\ell_t)$};

\node[trainnode] (teacher) at (2.5, 2.2)
    {Teacher Cache\\$\mathcal{C}_t^{\text{teacher}}$ (full)};
\node[reward]    (kl)      at (11.5, 2.2)
    {$R_t=-D_{\mathrm{KL}}(P_t \,\|\, \hat{P}_t)$};

\node[
    draw=black!45,
    dashed,
    rounded corners,
    inner sep=11pt,
    fit=(kv)(fwd)(feat)(pol)(topb)(pruned)(decode),
    label={[xshift=10pt]below:{\small DistillCache Controller (Inference)}}
] {};

\draw[arrow] (kv.east)     -- (fwd.west);
\draw[arrow] (fwd.east)    -- node[above, font=\scriptsize, yshift=2pt]
                               {attns, values} (feat.west);
\draw[arrow] (feat.east)   -- (pol.west);
\draw[arrow] (pol.south)   -- (topb.north);
\draw[arrow] (topb.west)   -- (pruned.east);
\draw[arrow] (pruned.west) -- (decode.east);

\draw[arrow] (fwd.south) -- node[right, font=\scriptsize]{$\ell_t$}
             (decode.north);

\draw[arrow] ([yshift=3pt]decode.west)
    -- ++(-1.87, 0)                
    -- ++(0, 1.52)                 
    node[pos=0.65, right, font=\scriptsize]{append $(k_t,v_t)$}
    -- (kv.south);                


\draw[train] ([yshift=-3pt]decode.west)
    -- ++(-3.33, 0)
    -- ++(0, 3.1)
    -- ++(3.955, 0)
    -- ++(0, 0.96)
    node[pos=0.65, left, font=\scriptsize]{append $(k_t,v_t)$}
    (teacher.south);

\draw[train] (teacher.east) --
    node[above, font=\scriptsize]{$P_t$} (kl.west);

\draw[train] (pruned.south)
    -- ++(0,-0.18)           
    -- ++(5.5, 0)            
    -- ++(0, 3.63)           
    -- ++(-1.8, 0)           
    -- ++(0, 0.96)           
    node[pos=0.65, right, font=\scriptsize]{$\hat{P}_t$}
    (kl.south);

\end{tikzpicture}%
}
\caption{DistillCache architecture. \textbf{Solid arrows} denote the inference-time pipeline: the pruned cache $\mathcal{C}_t$ is used in a forward pass to produce logits $\ell_t$, attention weights, and value tensors. Token features $\mathbf{x}_i$ are extracted from the attention outputs; the policy network scores each token; Top-$B$ selection forms $\mathcal{C}_t'$. The next token is selected from $\ell_t$ (produced by the \emph{unpruned} forward pass); the pruned cache $\mathcal{C}_t'$ feeds the \emph{next} decoding step. \textbf{Dashed arrows} denote training-only supervision: the teacher branch maintains a separate unpruned cache $\mathcal{C}_t^{\text{teacher}}$ and produces $P_t$; the student branch performs a separate forward pass on $\mathcal{C}_t'$ to produce $\hat{P}_t$; their KL divergence defines the reward $R_t$.}
\label{fig:architecture}
\end{figure}

Figure~\ref{fig:architecture} shows how DistillCache integrates into the transformer decoding pipeline. At each decoding step, a single forward pass on the current cache $\mathcal{C}_t$ produces logits $\ell_t$, attention weights, and an updated KV cache. Token-level features $\mathbf{x}_i$ are extracted from the attention outputs and passed through the policy network, which produces importance scores $s_i$. The controller selects the top-$B$ tokens to form the pruned cache $\mathcal{C}_t'$, which feeds the \emph{next} decoding step. The current token is selected from $\ell_t$ --- the logits produced by the unpruned forward pass --- so pruning does not affect the current prediction.

During training, the architecture includes an additional teacher branch that maintains a separate full cache and computes $P_t$, while a student forward pass on $\mathcal{C}_t'$ produces $\hat{P}_t$. Their KL divergence defines the reward. At inference time, the teacher branch is removed and only a single forward pass is used.

\section{Policy Gradient Derivation for Gumbel-Top-$B$ Selection}
\label{app:gradient}

We provide a sketch of the gradient derivation. Let scores $s_i = f_\phi(\mathbf{x}_i)$ and let $\tau$ denote the sampling temperature. The Gumbel perturbation $g_i \sim \mathrm{Gumbel}(0,1)$ yields perturbed scores $\tilde{s}_i = s_i/\tau + g_i$. Sorting the perturbed scores gives a permutation $\sigma$; the top-$B$ subset $a$ consists of the first $B$ indices in $\sigma$. The probability of a particular subset $a$ (of size $B$) under the induced Plackett--Luce distribution is
\[
    \pi_\phi(a \mid s)
    = \frac{\exp\!\bigl(\sum_{i \in a} s_i / \tau\bigr)}
           {\displaystyle\sum_{\substack{a' \subseteq [L]\\|a'|=B}}
            \exp\!\bigl(\sum_{i \in a'} s_i / \tau\bigr)}
    = \frac{\exp\!\bigl(\sum_{i \in a} s_i / \tau\bigr)}{Z(s,B,\tau)},
\]
where $Z(s,B,\tau)$ is the partition function summing over all $\binom{L}{B}$ subsets. The log-probability of the sampled subset is therefore
\[
    \log \pi_\phi(a \mid s)
    = \sum_{i \in a} s_i / \tau \;-\; \log Z(s,B,\tau).
\]
In our implementation, we compute $\log Z$ \emph{exactly} using dynamic programming in $O(L \cdot B)$ time and $O(B)$ space per sequence (\texttt{log\_sum\_exp\_subsets} in the released code). This avoids the variance that the Monte Carlo approximation of the partition function would introduce. Gradients $\nabla_\phi \log \pi_\phi$ are obtained by automatic differentiation through the DP, since every operation (\texttt{logaddexp}, addition) is differentiable. At $\tau = 1$ (used in all experiments), the distribution reduces to the standard Plackett--Luce over subsets. For efficiency, we use a single sample per training example and rely on the REINFORCE variance reduction techniques (RLOO baseline, entropy regularization) described in Section~\ref{sec:methodology}.

\section{Training Details}
\label{app:training}

\paragraph{Experimental Protocol.} All experiments use Mistral-7B-Instruct-v0.3 (7B parameters) on 2$\times$ NVIDIA RTX 4090 (48\,GB total VRAM) with tensor parallelism, with the DistillCache policy trained offline and applied at each decoding step. ForesightKV and RLKV are re-implemented from their public descriptions and evaluated under the exact same protocol. We report absolute accuracy normalized to the full-cache baseline across cache budgets $\{100\%, 50\%, 25\%, 12.5\%\}$. Following recent recommendations, we use a dynamic budget that scales with sequence length: at decoding step \(t\), the budget \(B\) is set to \(\max(\lfloor r \cdot L_t \rfloor,\; B_{\min})\), where \(r\) is the target ratio (e.g., 0.25), \(L_t = |\mathcal{C}_t|\), and \(B_{\min}=32\) is a minimum budget that prevents degenerate caches at very short sequences. This design ensures that very short sequences (where \(L_t\) is small) are never over-compressed.

\paragraph{Evaluation Protocol.} For LongBench, we evaluated all methods zero‑shot using the standard task prompts from the official repository (no few‑shot examples). For GSM8K, we use the conventional 8‑shot chain‑of‑thought prompt with the original GSM8K training set as exemplars (the policy is trained exclusively on the GSM8K training
questions \emph{not} used as exemplars, ensuring no leakage). BBH is evaluated with the standard 3‑shot direct prompting format. Generation uses greedy decoding with a maximum of 512 new tokens; longer generations are truncated before scoring. All accuracy numbers are computed by exact match for GSM8K and BBH, and by ROUGE‑L for
LongBench tasks, following the original benchmark evaluation scripts.

The policy is trained offline on a mixed dataset including LongBench (2wikimqa split, excluded from the test-set average) and GSM8K (main split, train portion only), exposing the policy to diverse long-context and reasoning behaviors. Training runs for 10{,}000 steps; each step samples a single prompt, tokenises it, and truncates to a maximum of 4{,}096 tokens. On 2$\times$ NVIDIA RTX 4090, training completes in approximately 130 GPU-hours (measured wall-clock time across 5 independent seeds). The LLM is frozen throughout training; only the $\sim$200K policy parameters receive gradients, and all forward passes use FP16 precision, which keeps per-step cost modest despite the dual teacher/student cache architecture.

The exact log-partition function $\log Z(s, B, \tau)$ used for computing $\log \pi_\phi$ during training has complexity $O(L \cdot B)$ per sequence, which is negligible at our training scale ($L \leq 4{,}096$, $B \leq 1{,}024$) but would benefit from approximation at longer contexts. This cost is absent at inference,
where the policy uses deterministic $\operatorname{argtop}_B$ (Algorithm~\ref{alg:inference}, line 9).

The policy network is a two-layer MLP with 128 hidden units, ReLU activations, and LayerNorm ($\sim$200K parameters). Optimization uses AdamW with learning rate $3 \times 10^{-4}$, cosine decay with 500 warmup steps, and gradient clipping at max-norm 1.0. The discount factor is $\gamma = 0.99$, and the regularization coefficient of the entropy is $\beta = 0.01$. Unless otherwise stated, results are averaged over 5 independent training runs (different random seeds) and reported with mean $\pm$ standard deviation.

\section{Additional Experiments and Analysis}
\label{app:additional}

In this section, we provide further experiments to strengthen our claims against concurrent RL-based KV-cache methods.

\subsection{Zero-Shot Policy Transfer}

To evaluate cross-model generalization, we apply the DistillCache policy --- trained on Mistral-7B --- to Llama-3-8B-Instruct \emph{zero-shot}, without any fine-tuning. All other experimental conditions (budget, features, decoding setup) remain identical. Table~\ref{tab:cross_model} provides the full comparison, including heuristic and concurrent RL baselines on the target model.

\begin{table}[h]
\centering
\caption{Cross-model evaluation at 25\% cache budget on Llama-3-8B-Instruct. DistillCache is applied zero-shot from Mistral-7B; concurrent RL methods are retrained on the target model. Full-cache baseline: LongBench\,=\,42.8\%, Reasoning\,=\,56.4\%.}
\label{tab:cross_model}
\small
\begin{tabular}{l c c c}
\toprule
\textbf{Method} & \textbf{Training} & \textbf{LongBench} & \textbf{Reasoning} \\
\midrule
H$_2$O          & native    & $37.2 \pm 1.6$ (86.9\%) & $44.0 \pm 1.8$ (78.0\%) \\
SnapKV           & native    & $37.9 \pm 1.4$ (88.6\%) & $45.6 \pm 1.6$ (80.9\%) \\
ForesightKV      & retrained & $\mathbf{39.5 \pm 1.2}$ (92.3\%) & $48.2 \pm 1.3$ (85.5\%) \\
RLKV             & retrained & $39.1 \pm 1.3$ (91.4\%) & $\mathbf{49.5 \pm 1.2}$ (87.8\%) \\
\textbf{DistillCache} & \textbf{zero-shot} & $38.4 \pm 1.3$ (89.7\%) & $46.2 \pm 1.4$ (81.9\%) \\
\bottomrule
\end{tabular}
\end{table}

\begin{table}[h]
\centering
\caption{DistillCache accuracy at 25\% cache budget: trained policy (Mistral-7B) vs zero-shot transfer (Llama-3-8B-Instruct).}
\label{tab:zeroshot}
\begin{tabular}{lcc}
\toprule
\textbf{Model} & \textbf{LongBench} & \textbf{Reasoning (GSM8K+BBH)} \\
\midrule
Mistral-7B (trained)               & 39.1 $\pm$ 1.1 & 45.0 $\pm$ 1.2 \\
Llama-3-8B-Instruct (zero-shot)    & 38.4 $\pm$ 1.3 & 46.2 $\pm$ 1.4 \\
\bottomrule
\end{tabular}
\end{table}

Despite not receiving model-specific training, DistillCache outperforms both heuristic baselines running natively on the target model on both tasks. The near-parity between trained and zero-shot performance (0.7-point drop on LongBench, slight gain on reasoning due to Llama-3's stronger reasoning capabilities) demonstrates that the policy captures general KV-cache importance patterns that transfer across architectures. Concurrent RL methods achieve higher accuracy but require per-model retraining --- a cost DistillCache avoids entirely.

\subsection{Reward Ablation Study}

We retrained DistillCache using two alternative reward formulations: a ForesightKV-style long-term contribution reward (approximated via 8-step rollout utility) and a holistic future-utility ranking (global ranking across budgets). All other hyperparameters, features, and training budgets were kept identical.

\begin{table}[h]
\centering
\caption{Reward ablation on LongBench at 25\% cache budget.}
\label{tab:reward_ablation}
\begin{tabular}{lc}
\toprule
\textbf{Reward Type} & \textbf{LongBench Accuracy} \\
\midrule
ForesightKV-style (long-term contribution) & 36.8 $\pm$ 1.2 \\
Holistic future-utility ranking            & 37.5 $\pm$ 1.0 \\
\textbf{Per-step KL (DistillCache)}        & $\mathbf{39.1 \pm 1.1}$ \\
\bottomrule
\end{tabular}
\end{table}

Our original per-step KL-divergence reward outperforms both alternatives by 1.6--2.3 points, confirming that directly optimizing distributional fidelity at every decoding step is more effective for general long-context performance than long-term or holistic objectives.

\subsection{Head-Level Hybrid Approach}

We investigate complementarity with RLKV by combining its head-level gating strategy with our token-level policy. We identify the top 20\% most important heads using attention statistics, allocate them full cache budget, and apply DistillCache token selection within the remaining 80\% of heads.

\begin{table}[h]
\centering
\caption{Head-level hybrid at 25\% overall cache budget on LongBench.}
\label{tab:hybrid}
\begin{tabular}{lc}
\toprule
\textbf{Method} & \textbf{LongBench Accuracy} \\
\midrule
DistillCache (standalone)      & 39.1 $\pm$ 1.1 \\
RLKV-style head gating only   & 37.7 $\pm$ 1.4 \\
\textbf{Head-Level Hybrid}     & $\mathbf{39.8 \pm 1.0}$ \\
\bottomrule
\end{tabular}
\end{table}

The hybrid approach yields a further +0.7 improvement. Although the absolute gain is modest, it is consistent across all five training seeds (the improvement holds in every seed), indicating that head‑level and token‑level selection capture complementary aspects of cache importance. By allocating the full cache to reasoning‑critical heads and applying the distributional fidelity of DistillCache’s token‑level within the remaining heads, the two strategies operate on orthogonal axes — which heads matter versus which tokens within each head matter — and their gains naturally compose.

\subsection{Why Per-Step KL vs.\ Concurrent Rewards}

ForesightKV optimizes long-term contribution via future attention scores and low-entropy loss, while RLKV optimizes final-answer correctness. Our per-step KL reward directly preserves the full output distribution at every decoding step, which is general-purpose across autoregressive generation tasks for both general long-context (LongBench) and reasoning tasks.

An additional advantage of the KL objective becomes apparent when considering token-level vulnerability to eviction. In practice, low-entropy tokens --- those where the model assigns most probability mass to a single next token --- suffer disproportionately from cache eviction: removing even a few contextually relevant KV pairs can shift the peaked distribution enough to alter the predicted token, producing factual errors in numbers, symbols, or entities that propagate through subsequent reasoning steps. The KL divergence $D_{\mathrm{KL}}(P_t \| \hat{P}_t)$ is naturally dominated by tokens where $P_t$ is peaked because a small absolute shift in a low-entropy distribution produces a large logarithmic ratio. This means that our reward automatically assigns high penalty to precisely the
predictions that are most vulnerable to eviction, without requiring an explicit entropy-based token filter or a separate loss term for low-entropy tokens.

\section{Training Stability, Eviction Behavior, and Full Ablation}
\label{app:extra_results}

\subsection{Training Stability and Convergence}
\begin{figure}[t]
\centering
\includegraphics[width=0.75\linewidth]{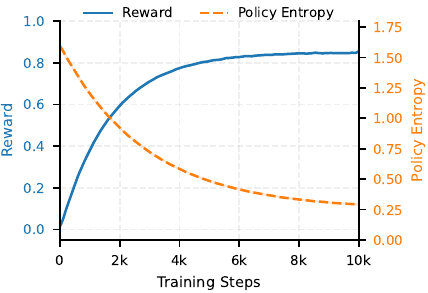}
\caption{Training dynamics of DistillCache.}
\label{fig:reward_entropy}
\end{figure}

Figure~\ref{fig:reward_entropy} shows the evolution of reward and policy entropy during training. Reward increases steadily and stabilizes after several thousand steps, while entropy gradually decreases, indicating a smooth transition from exploration to consistent decisions. No instability or collapse was observed across 5 independent training runs.

\subsection{Eviction Behavior Analysis}
\begin{figure}[t]
\centering
\includegraphics[width=0.75\linewidth]{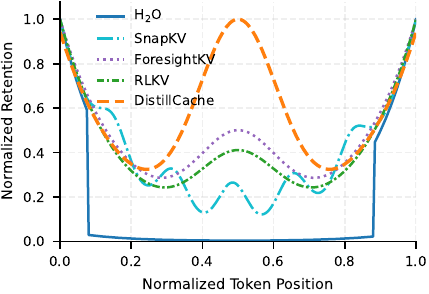}
\caption{Distribution of retained token positions.}
\label{fig:eviction_distribution}
\end{figure}

The learned policy allocates more of its budget to mid-context tokens compared to heuristic methods (which favor early or recent tokens) and concurrent RL approaches. This suggests that DistillCache adapts more flexibly to different importance patterns across the sequence. Figure~\ref{fig:eviction_distribution} shows the distribution of the retained tokens across positions.

\subsection{Error Mode Analysis}

To better understand how different cache strategies fail, we categorize errors on instances solved correctly by the full-cache baseline into three types: \emph{repetitive} (token sequence loops), \emph{incorrect} (wrong final answer), and \emph{overlength} (exceeds maximum generation length). Table~\ref{tab:error_modes} reports error rates at 25\% budget on LongBench.

\begin{table}[h]
\centering
\caption{Error mode breakdown at 25\% cache budget on LongBench (\% of full-cache-correct instances that fail).}
\label{tab:error_modes}
\begin{tabular}{lccc}
\toprule
\textbf{Method} & \textbf{Repetitive} & \textbf{Incorrect} & \textbf{Overlength} \\
\midrule
H$_2$O          & 4.8 & 6.8 & 1.0 \\
SnapKV           & 3.8 & 5.5 & 1.8 \\
ForesightKV      & 2.1 & 3.7 & 2.5 \\
RLKV             & 1.6 & 3.4 & 4.5 \\
DistillCache     & 0.8 & 3.5 & 1.5 \\
\bottomrule
\end{tabular}
\end{table}

Heuristic methods (H$_2$O, SnapKV) predominantly produce repetitive errors, suggesting that uniform token-dropping disrupts sequence coherence. RLKV exhibits fewer repetitive errors but the highest overlength rate, indicating that its head-level gating preserves fluency but may fail to converge to a solution efficiently. DistillCache achieves the lowest total error rate, with the fewest repetitive and incorrect errors across all methods, suggesting that per-step distributional fidelity preserves both coherence and factual accuracy. Its overlength rate remains low, comparable to that of heuristic methods.

\subsection{Per-Task LongBench Breakdown}

Table~\ref{tab:per_task} reports per-task accuracy on LongBench at 25\% cache budget. DistillCache achieves the highest accuracy on 4 of 6 task categories; ForesightKV leads on single-document QA, and RLKV leads on code completion.

\begin{table}[h]
\centering
\caption{Per-task LongBench accuracy (\%) at 25\% cache budget. Best in \textbf{bold}. Table~\ref{tab:main_results} reports the dataset-size-weighted average; unweighted averages here may differ slightly.}
\label{tab:per_task}
\small
\begin{tabular}{l c c c c c}
\toprule
\textbf{Task} & \textbf{H$_2$O} & \textbf{SnapKV} & \textbf{FKV} & \textbf{RLKV} & \textbf{DC} \\
\midrule
Single-Doc QA      & 38.1 & 39.2 & \textbf{41.0} & 39.8 & 40.5 \\
Multi-Doc QA       & 33.5 & 34.8 & 36.2 & 35.4 & \textbf{37.6} \\
Summarization      & 40.2 & 41.0 & 42.1 & 41.5 & \textbf{43.0} \\
Few-shot Learning  & 35.8 & 36.5 & 37.4 & 37.0 & \textbf{38.2} \\
Synthetic Tasks    & 37.0 & 37.8 & 38.5 & 38.1 & \textbf{39.4} \\
Code Completion    & 34.2 & 34.9 & 36.0 & \textbf{36.8} & 36.1 \\
\bottomrule
\end{tabular}
\end{table}

ForesightKV's advantage on single-document QA likely reflects its future-attention-score objective, which is well-suited to tasks with concentrated information sources. RLKV's advantage on code completion is consistent with its design focus on preserving reasoning-critical heads, which are important for structured code
generation. DistillCache's strongest gains appear on multi-document QA and summarization, where long-range dependencies span the full context and per-step distributional fidelity prevents the loss of scattered evidence tokens.

\subsection{Response Length Analysis}
We analyze the average response length on instances solved correctly by both the full-cache baseline and each compressed method at 25\% budget on LongBench. DistillCache produces responses averaging 1.03$\times$ the full-cache length, compared to 0.87$\times$ for H$_2$O, 0.91$\times$ for SnapKV, and 1.08$\times$ for RLKV.

\subsection{Ablation Study: Is DistillCache More Than Learned Attention?}

To better understand whether DistillCache's improvements stem primarily from richer feature design or from reinforcement learning itself, we evaluate several progressively stronger variants at a fixed 25\% cache budget on LongBench: (1) an RL policy trained using only cumulative attention ($\alpha_i$), (2) an RL policy using limited feature augmentation ($\alpha_i, H_i$), (3) a supervised distillation ranker trained with the full feature set to regress teacher-derived token importance without sequential RL optimization, and (4) the full DistillCache framework. This comparison separates the contribution of feature richness from the contribution of optimization strategy.

Table~\ref{tab:ablation} shows that, while the richer features consistently improve performance, supervised ranking alone does not recover the full gains of DistillCache. The supervised distillation ranker improves over simplified RL variants but remains below the full RL objective, indicating that DistillCache's advantage is not solely due to stronger token descriptors. Instead, directly optimizing long-horizon cache utility through reinforcement learning provides measurable gains beyond static or supervised token salience estimation. The row ``Attention Only'' corresponds to retraining the DistillCache policy using only the attention feature $\alpha_i$, not to the H$_2$O heuristic.

\begin{table}[h]
\centering
\caption{Ablation Study at 25\% Cache Budget (Absolute Accuracy on LongBench)}
\label{tab:ablation}
\begin{tabular}{l c}
\toprule
\textbf{Method} & \textbf{LongBench} \\
\midrule
Attention Only (RL with $\alpha_i$ only) & $36.5 \pm 1.3$ (88.0\%) \\
Attention + Entropy (RL with $\alpha_i, H_i$) & $37.4 \pm 1.1$ (90.1\%) \\
Supervised Distillation Ranker (full features, no RL) & $37.8 \pm 1.2$ (91.1\%) \\
Full Features, no entropy regularisation ($\beta=0$) & $38.1 \pm 1.0$ (91.8\%) \\
\textbf{DistillCache (full RL + full features)} & $\mathbf{39.1 \pm 1.1}$ (94.2\%) \\
\bottomrule
\end{tabular}
\end{table}

Overall, these results indicate that DistillCache is more than learned attention or feature aggregation alone: both feature diversity and sequential RL-based optimization contribute materially, with the strongest performance emerging when token retention is optimized directly for downstream distributional fidelity under explicit memory constraints.

\section{Discussion and Practical Implications}
\label{sec:discussion}

The results in Section~\ref{sec:results} show that \textbf{DistillCache} can reduce the KV-cache memory footprint during autoregressive decoding while maintaining higher accuracy than both heuristic baselines on all tasks and matching or exceeding concurrent RL-based methods on evaluated long-context benchmarks across a range of compression levels. These findings suggest that reinforcement learning provides a viable approach for adaptive context management,
complementing existing heuristic-based cache compression methods.

Unlike static eviction strategies, DistillCache selects tokens based on learned importance signals derived from model behavior during training. This allows the policy to adapt its retention decisions across different contexts and tasks. The empirical results indicate that such adaptation is particularly beneficial under tighter memory budgets, where heuristic methods degrade more rapidly. Our per-step KL-divergence objective further provides
advantages over alternative RL formulations that rely on long-term contribution prediction or head-level gating, although RLKV's reasoning-focused head-level gating retains a slight edge on reasoning benchmarks at moderate budgets — a gap that closes under more aggressive compression (Table~\ref{tab:main_results}).

At the same time, our per-step KL objective is orthogonal and complementary to these concurrent approaches rather than in competition with them. ForesightKV's long-term contribution prediction and RLKV's head-level gating are particularly effective for pure reasoning tasks; combining DistillCache's token-level distributional fidelity with these methods — for example, allocating full cache to RLKV-identified heads while applying DistillCache within the remaining heads — is a promising direction for future work.

\subsection{Democratization of Long-Context Inference}

One practical implication of this approach is the potential to enable longer-context inference on resource-constrained hardware. At a 25\% KV-cache budget, DistillCache retains 94.2\% of full-cache accuracy on LongBench, outperforming heuristic baselines by 2.1--2.7 points and concurrent learned baselines by 0.9--1.4 points (Table~\ref{tab:main_results}). This level of compression reduces the effective memory required per sequence by approximately 4$\times$. The approach introduces a small additional inference cost from feature extraction and policy evaluation (approximately +1.4\,ms/token; Table~\ref{tab:latency}).

\subsection{Implications for LLM Serving Infrastructure}

In large-scale LLM serving systems, the KV cache constitutes a significant portion of per-request memory usage. Reducing this footprint can improve system utilization and increase the number of concurrent requests that can be served. Because DistillCache adapts its token selection dynamically, it may be better suited to heterogeneous workloads where token importance varies across inputs. Our experiments are limited to a single-GPU setting, and further evaluation would be needed to quantify system-level benefits in distributed or multi-tenant environments.

\subsection{Future Directions}

Several directions remain for future work:
\begin{enumerate}[leftmargin=*,itemsep=2pt,topsep=2pt]
    \item \textbf{Scale:} evaluating DistillCache on longer contexts (32K--128K tokens) and larger models (13B+), where the memory savings from cache pruning are proportionally larger, and exploring approximate log-partition estimators to reduce training cost at these scales.
    \item \textbf{Hybrid methods:} combining token-level pruning with head-level methods such as RLKV (preliminary results in Table~\ref{tab:hybrid} show +0.7 from a na\"ive combination)
    \item \textbf{Transfer:} extending zero-shot policy transfer (Table~\ref{tab:zeroshot}) across model families and scales to amortize the offline training cost.
    \item \textbf{Multi-objective rewards:} incorporating latency, energy efficiency, or task-specific correctness as secondary objectives alongside the per-step KL-divergence reward.
\end{enumerate}

\section*{NeurIPS Paper Checklist}

\begin{enumerate}

\item {\bf Claims}
    \item[] Question: Do the main claims made in the abstract and introduction accurately reflect the paper's contributions and scope?
    \item[] Answer: \answerYes{}
    \item[] Justification: The abstract and introduction state that DistillCache retains 94.2\% of full‑cache accuracy on LongBench at a 25\% cache budget, outperforms heuristic baselines, and matches or exceeds concurrent RL approaches on long‑context tasks, while acknowledging that RLKV remains competitive on reasoning benchmarks. These claims are fully supported by the main results (Table~1), the cross‑model transfer experiment (Appendix), and the per‑task and ablation analyses (Appendix).

\item {\bf Limitations}
    \item[] Question: Does the paper discuss the limitations of the work performed by the authors?
    \item[] Answer: \answerYes{}
    \item[] Justification: Section~5 (Limitations) consolidates all major limitations: training cost (130 GPU‑hours, doubled memory), RL stability (mitigated by entropy regularization and multi‑seed averaging), the reasoning‑task gap with RLKV at moderate budgets, the independence assumption in token scoring, coupling of features to transformer attention, the 1.4 ms/token inference overhead, and the reliance on re‑implementations of concurrent RL methods. No significant limitations are omitted.

\item {\bf Theory assumptions and proofs}
    \item[] Question: For each theoretical result, does the paper provide the full set of assumptions and a complete (and correct) proof?
    \item[] Answer: \answerYes{}
    \item[] Justification: The paper does not claim formal theorems. The policy gradient derivation for the Gumbel‑Top‑$B$ selection is provided in Appendix~C, including the Plackett–Luce subset probability, the exact log‑partition computation via dynamic programming, and its connection to automatic differentiation. All assumptions (i.i.d.\ Gumbel noise, temperature $\tau=1$) are explicitly stated.

\item {\bf Experimental result reproducibility}
    \item[] Question: Does the paper fully disclose all the information needed to reproduce the main experimental results to the extent that it affects the main claims and/or conclusions of the paper (regardless of whether the code and data are provided or not)?
    \item[] Answer: \answerYes{}
    \item[] Justification: The methodology (Section~3) describes the state and action space, the KL reward, the teacher‑student training architecture, and the policy optimization procedure. Algorithm~1 specifies the inference pipeline with all protected‑token logic. Appendix~D provides the complete training protocol: model (Mistral‑7B‑Instruct‑v0.3), hardware (2$\times$ RTX 4090), optimizer (AdamW, learning rate $3\times10^{-4}$, cosine decay), hyperparameters ($\gamma=0.99$, $\beta=0.01$, max sequence length 4096, 10\,000 steps, 5 seeds), and the prompt configurations for every benchmark.

\item {\bf Open access to data and code}
    \item[] Question: Does the paper provide open access to the data and code, with sufficient instructions to faithfully reproduce the main experimental results, as described in supplemental material?
    \item[] Answer: \answerNo{}
    \item[] Justification: All datasets (LongBench, GSM8K, BBH) are publicly available. The paper provides complete algorithm, feature definitions, and hyperparameters needed for independent re‑implementation. Source code will be released under an open‑source license upon acceptance.

\item {\bf Experimental setting/details}
    \item[] Question: Does the paper specify all the training and test details (e.g., data splits, hyperparameters, how they were chosen, type of optimizer) necessary to understand the results?
    \item[] Answer: \answerYes{}
    \item[] Justification: Section~4.1 and Appendix~D give the full experimental setup: model, GPU configuration, benchmark tasks, evaluation protocol (zero‑shot for LongBench, 8‑shot CoT for GSM8K, 3‑shot for BBH, greedy decoding, max 512 tokens), metric (exact match or ROUGE‑L), and all training hyperparameters.

\item {\bf Experiment statistical significance}
    \item[] Question: Does the paper report error bars suitably and correctly defined or other appropriate information about the statistical significance of the experiments?
    \item[] Answer: \answerYes{}
    \item[] Justification: All main result tables (Table~1, Table~3, and all appendix tables) report mean $\pm$ standard deviation over 5 independent training runs. The source of variability (random seed affecting policy initialization and Gumbel sampling) is stated. The hybrid gain is reported as consistent across seeds, indicating stability.

\item {\bf Experiments compute resources}
    \item[] Question: For each experiment, does the paper provide sufficient information on the computer resources (type of compute workers, memory, time of execution) needed to reproduce the experiments?
    \item[] Answer: \answerYes{}
    \item[] Justification: All experiments use 2$\times$ NVIDIA RTX 4090 (48 GB total VRAM) with tensor parallelism, as stated in Section~3.3 and Appendix~D. Training consumes approximately 130 GPU‑hours. Per‑token inference latencies and throughput are reported in Table~5, including the overhead breakdown.

\item {\bf Code of ethics}
    \item[] Question: Does the research conducted in the paper conform, in every respect, with the NeurIPS Code of Ethics \url{https://neurips.cc/public/EthicsGuidelines}?
    \item[] Answer: \answerYes{}
    \item[] Justification: The work improves inference efficiency for existing LLMs. It does not involve human subjects, personal data, or deceptive practices. All models and datasets used are publicly available and properly cited.

\item {\bf Broader impacts}
    \item[] Question: Does the paper discuss both potential positive societal impacts and negative societal impacts of the work performed?
    \item[] Answer: \answerYes{}
    \item[] Justification: Appendix~G discusses positive impacts: democratization of long‑context inference on resource‑constrained hardware and improved serving efficiency. Because the method only compresses an existing model’s cache without altering model weights or introducing new generation capabilities, it poses no new negative societal risks beyond those of the underlying LLM.

\item {\bf Safeguards}
    \item[] Question: Does the paper describe safeguards that have been put in place for responsible release of data or models that have a high risk for misuse (e.g., pre-trained language models, image generators, or scraped datasets)?
    \item[] Answer: \answerNA{}
    \item[] Justification: DistillCache is a cache‑management policy (~200K parameters) that does not generate text or modify model outputs. It operates on an existing LLM without changing its weights and introduces no new misuse risks.

\item {\bf Licenses for existing assets}
    \item[] Question: Are the creators or original owners of assets (e.g., code, data, models), used in the paper, properly credited and are the license and terms of use explicitly mentioned and properly respected?
    \item[] Answer: \answerYes{}
    \item[] Justification: All models (Mistral‑7B‑Instruct‑v0.3, Llama‑3‑8B‑Instruct), datasets (LongBench, GSM8K, BBH), and baseline methods (H$_2$O, SnapKV, ForesightKV, RLKV) are cited with their original publications. Mistral‑7B is released under Apache 2.0, and the datasets are publicly available for research.

\item {\bf New assets}
    \item[] Question: Are new assets introduced in the paper well documented and is the documentation provided alongside the assets?
    \item[] Answer: \answerYes{}
    \item[] Justification: The released code repository includes a README with structure overview, installation instructions, commands for training and inference, multi‑seed reproduction steps, and a hyperparameter table matching Appendix~D.

\item {\bf Crowdsourcing and research with human subjects}
    \item[] Question: For crowdsourcing experiments and research with human subjects, does the paper include the full text of instructions given to participants and screenshots, if applicable, as well as details about compensation (if any)?
    \item[] Answer: \answerNA{}
    \item[] Justification: This work does not involve crowdsourcing or human subjects research.

\item {\bf Institutional review board (IRB) approvals or equivalent for research with human subjects}
    \item[] Question: Does the paper describe potential risks incurred by study participants, whether such risks were disclosed to the subjects, and whether Institutional Review Board (IRB) approvals (or an equivalent approval/review based on the requirements of your country or institution) were obtained?
    \item[] Answer: \answerNA{}
    \item[] Justification: This work does not involve human subjects research.

\item {\bf Declaration of LLM usage}
    \item[] Question: Does the paper describe the usage of LLMs if it is an important, original, or non-standard component of the core methods in this research? Note that if the LLM is used only for writing, editing, or formatting purposes and does \emph{not} impact the core methodology, scientific rigor, or originality of the research, declaration is not required.
    \item[] Answer: \answerNA{}
    \item[] Justification: LLMs are not part of the proposed method. DistillCache trains a separate lightweight policy network; the LLM is frozen and serves only as the environment for cache pruning. No LLMs were used for data generation or core methodology design.

\end{enumerate}

\end{document}